\documentclass[sigconf,nonacm]{aamas} 

\usepackage{balance} 

\usepackage{tikz}
\usetikzlibrary{automata, positioning, arrows, calc}
\usepackage{tabularx}
\usepackage{hyperref}

\usepackage[title]{appendix}
\usepackage[capitalise,noabbrev]{cleveref}

\setcopyright{ifaamas}
\acmConference[AAMAS '23]{Proc.\@ of the 22nd International Conference
on Autonomous Agents and Multiagent Systems (AAMAS 2023)}{May 29 -- June 2, 2023}
{London, United Kingdom}{A.~Ricci, W.~Yeoh, N.~Agmon, B.~An (eds.)}
\copyrightyear{2023}
\acmYear{2023}
\acmDOI{}
\acmPrice{}
\acmISBN{}

\acmSubmissionID{???}

\title[The SMAClite environment]{SMAClite: A Lightweight Environment for Multi-Agent Reinforcement Learning}

\author{Adam Michalski}
\affiliation{
  \institution{University of Edinburgh}
  \city{Edinburgh}
  \country{United Kingdom}}
\email{contact@adammi.ch}

\author{Filippos Christianos}
\affiliation{
  \institution{University of Edinburgh}
  \city{Edinburgh}
  \country{United Kingdom}}
\email{f.christianos@ed.ac.uk}

\author{Stefano V. Albrecht}
\affiliation{
  \institution{University of Edinburgh}
  \city{Edinburgh}
  \country{United Kingdom}}
\email{s.albrecht@ed.ac.uk}

\begin{abstract}
There is a lack of standard benchmarks for Multi-Agent Reinforcement Learning (MARL) algorithms. The Starcraft Multi-Agent Challenge (SMAC) has been widely used in MARL research, but is built on top of a heavy, closed-source computer game, StarCraft II. Thus, SMAC is computationally expensive and requires knowledge and the use of proprietary tools specific to the game for any meaningful alteration or contribution to the environment. We introduce SMAClite -- a challenge based on SMAC that is both decoupled from Starcraft II and open-source, along with a framework which makes it possible to create new content for SMAClite without any special knowledge. We conduct experiments to show that SMAClite is equivalent to SMAC, by training MARL algorithms on SMAClite and reproducing SMAC results. We then show that SMAClite outperforms SMAC in both runtime speed and memory.
\end{abstract}


\newcommand{\BibTeX}{\rm B\kern-.05em{\sc i\kern-.025em b}\kern-.08em\TeX}

\newcommand{\cN}{\mathcal{N}}
\newcommand{\cS}{\mathcal{S}}
\newcommand{\cA}{\mathcal{A}}
\newcommand{\cP}{\mathcal{P}}
\newcommand{\cR}{\mathcal{R}}

\newcommand{\cO}{\mathcal{O}}

\begin{document}


\pagestyle{fancy}
\fancyhead{}


\maketitle

\section{Introduction}
As we continue to make advancements in artificial intelligence research, it inevitably makes its way into our daily lives. Examples of autonomous agents popular in recent times range from robotic vacuum cleaners (e.g. \cite{vaussard_lessons_2014}) and self-driving cars (e.g. \cite{albrecht_interpretable_2021}) to robots making lives easier from behind the scenes, such as warehouse optimization robots (e.g. \cite{krnjaic_scalable_2022}). With all this attention in research, a natural need arises for standardized benchmarks for the various types of artificial intelligence (AI) models.

Multi-agent reinforcement learning (MARL) is a branch of machine learning dealing with multiple autonomous AI entities -- usually called \emph{agents} -- existing in the same environment. In this work, we are interested in \emph{cooperative} agents, ones that work together to accomplish some goal. Crucially, there is currently no consensus in the research community about what a standard benchmark for this type of agent should be. Just by looking at a recent benchmarking paper \cite{papoudakis_benchmarking_2021} we can count five different benchmarking environments.

The Starcraft Multi-Agent Challenge (SMAC) \cite{samvelyan_starcraft_2019} has been widely used in MARL research. It is built on top of a real-time strategy computer game Starcraft II (SC2) and makes use of an API -- an interface between the game and the AI agents -- made available by \citet{vinyals_starcraft_2017}. It presents a mini-game where each agent controls a single combat unit (e.g. a single soldier) in one of several available battle scenarios against an enemy team controlled by the game's built-in AI. SMAC presents a challenge where the solution is not straightforward -- in most of the scenarios the most obvious strategy of running forward and attacking is not good enough and will result in a quick loss due to the enemy army having better units or more numbers.

While the idea is promising, we can spot several problems with SMAC if it is to become a universally accepted benchmark. The biggest issue we see is that SMAC uses SC2 as its key dependency, requiring a large-sized (ca. 3.7 GB) download and a complicated setup process for any training or inference, not to mention running SC2 alongside SMAC at all times, consuming extra CPU and memory resources. This is made worse by the fact that SMAC uses only a subset of the SC2 features -- a lot of the required downloadable and computational resources are simply redundant, and due to that training agents on SMAC is more expensive than necessary for the task it offers.

On top of that, it remains highly inaccessible for people unfamiliar with the game it is based in. This manifests in many ways, e.g. to create custom scenarios or units for SMAC, one is required to use the official Starcraft II map editor, requiring people to learn an unusual and proprietary tool, and put in a lot of effort for a single benchmark.

We present a challenge very similar to SMAC, but completely decoupled from the Starcraft II dependency, and show that it preserves the challenging aspects of SMAC. We name this challenge Starcraft Multi-Agent Challenge lite (SMAClite). We also want to make the battle scenarios and units as easy to modify as possible -- also allowing easy creation of completely new ones. Our environment preserves the outer interface of SMAC, only changing the inner workings, in order to allow AI developers to reuse their code for handling SMAC with minimal modifications. We make this environment open-source\footnote{\href{https://github.com/uoe-agents/smaclite}{https://github.com/uoe-agents/smaclite}} and free to use.

We perform a series of experiments using models trained on SMAClite. Our experiments include quantitative analysis by comparing the return achieved by various MARL algorithms in SMAClite -- we show that the algorithms achieve similar returns as on SMAC and that the relative ranking among them is preserved from SMAC. We also perform qualitative analysis -- looking at the combat strategies employed by the agents on a case-by-case basis, and verifying they do indeed outsmart the handwritten enemy AI. On top of that, we take the models trained on SMAClite, and put them inside the original SMAC environment without any further training, to see how much potential for transfer learning there is between the environments -- from this experiment, we conclude that training agents on SMAClite improves the returns achieved by them on SMAC, and therefore the challenges require similar skills.

\section{Related Work}
Popular benchmarks, accepted by the research community as the standard, do exist for the single-agent variant of reinforcement learning \cite{plappert_multi-goal_2018, bellemare_arcade_2013}. The authors of the MARL benchmark paper \cite{papoudakis_benchmarking_2021} make available two MARL benchmarks, both based in simple 2D worlds: Level-Based Foraging, and Multi-Robot Warehouse. There are also several other non-Starcraft II multi-agent benchmarks, such as the Multi-Agent Particle Environment \cite{mordatch_emergence_2018}, or the Hanabi challenge \cite{bard_hanabi_2020}.  Another multi-agent challenge based on a modern computer game is the OpenAI Five project \cite{berner_dota_2019}, which put agents inside a full five versus five matches of the strategy game Dota 2, showing the impressive scale of the game and the trained models.

Training autonomous agents in Starcraft II started becoming popular upon the publication of its API \cite{vinyals_starcraft_2017}, with one of the popular results being the AlphaStar model \cite{vinyals_grandmaster_2019}. Our work is mostly based on the work of the creators of SMAC \cite{samvelyan_starcraft_2019}. Since its publication, SMAC has been used as a benchmark for autonomous agents in numerous works (e.g. \cite{papoudakis_benchmarking_2021, rashid_qmix_2018, yu_surprising_2021}), with its popularity being a sign that it does fill a niche.

There are several projects that are tangentially related to ours since they also aim to improve SMAC but take different approaches (e.g. SMACv2 \cite{ellis_smacv2_2022}, SMAC+ \cite{kim_starcraft_2022}). These projects present interesting additions to the SMAC paradigm, however, to our knowledge, none of them address the issues we wish to tackle: the environment's performance cost, and its closed-source nature. We believe the additions introduced by them are good ideas for the future development of SMAClite, but the scope of our project is to maintain the challenge's difficulty on the same level as the original.


\section{Background} \label{chap:background}
\subsection{Multi-Agent Reinforcement Learning} \label{sec:marl}

Multi-Agent Reinforcement learning (MARL) allows for multiple autonomous agents to coexist in the same space. Formally, the setup of MARL consists of several \emph{agents} that can perform specific actions, and the \emph{environment} -- a term that encompasses everything outside of the agents, that the agents can interact with. The specific formalisation of MARL that we use in our work is called \textbf{Decentralized Partially Observable Markov Decision Problems} (in short: Dec-POMDPs).

A Dec-POMDP is a cooperative process defined as a 7-element tuple $(\cN, \cS, \cA, \cO, \Omega, \cP, \cR)$, where $\cN = \{1\ldots N\}$ is the set of agents participating in the process. Agents interact with the environment in discrete timesteps $t \in \mathbf{N}$. In each timestep $t$ the environment has some true active state $s_t \in \cS$, and each agent $i$ receives an observation $o_t^i \sim \Omega(i, s_t),\ o_t^i \in \cO$. Each agent $i$ then selects an action $a^i_t \in \cA$. After each timestep $t$ the agents receive a shared reward $\cR(a^1_t, a^2_t, \ldots, a^N_t, s_t) = r_{t+1} \in \mathbf{R}$, and the environment enters the next state $s_{t+1} \sim \cP(a^1_t, a^2_t, \ldots, a^N_t, s_t),\ s_{t+1} \in \cS$.

In a Dec-POMDP, like in reinforcement learning in general, the agents' goal at each point in time $t$ is to maximize the discounted cumulative reward (or the \emph{return}) $\sum_{i=0}^\infty \gamma^i r_{t+i}$, where $\gamma$ is a discount factor. When $\gamma = 1$, the discounted return is equal to the actual return $\sum_t r_t$ -- we will report this sum when presenting evaluation results.

For the purpose of training and evaluating agents in the SMAClite environment, we will use the same set of algorithms, as well as their hyperparameters, that was used to test agents in SMAC in a recent MARL benchmark paper \cite{papoudakis_benchmarking_2021}. This includes 9 popular algorithms that can be used to solve various SMAC scenarios: IQL \cite{tan_multi-agent_1993}, IA2C \cite{mnih_asynchronous_2016}, IPPO \cite{schulman_proximal_2017}, MADDPG \cite{lowe_multi-agent_2017}, COMA \cite{foerster_counterfactual_2018}, MAA2C \cite{papoudakis_benchmarking_2021}, MAPPO \cite{yu_surprising_2021}, VDN \cite{sunehag_value-decomposition_2018}, and QMIX \cite{rashid_qmix_2018}.

\subsection{SMAC} \label{sec:smac}
\begin{figure*}[t]
    \centering
    \includegraphics[height=0.25\textwidth]{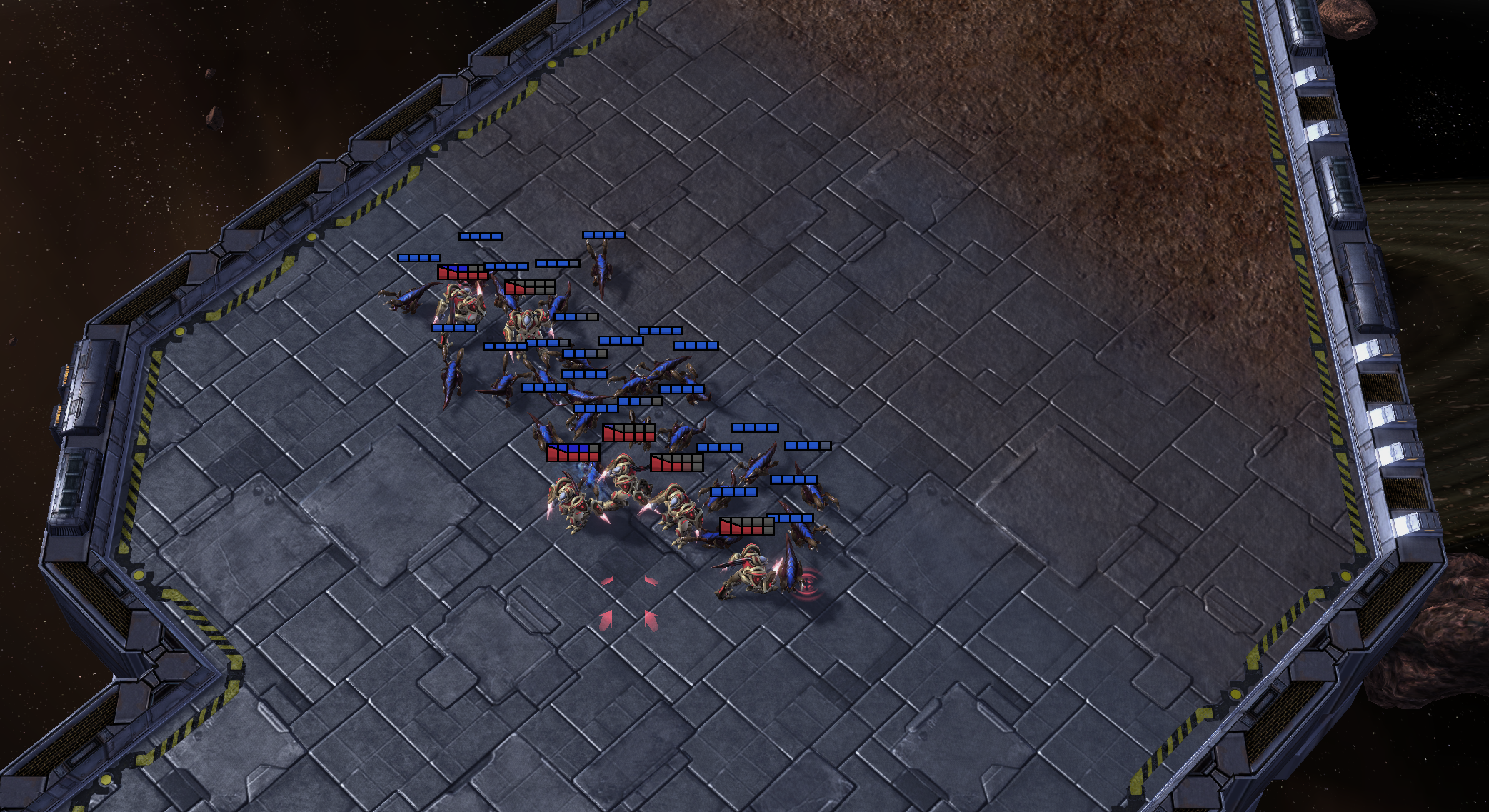}
    \includegraphics[height=0.25\textwidth]{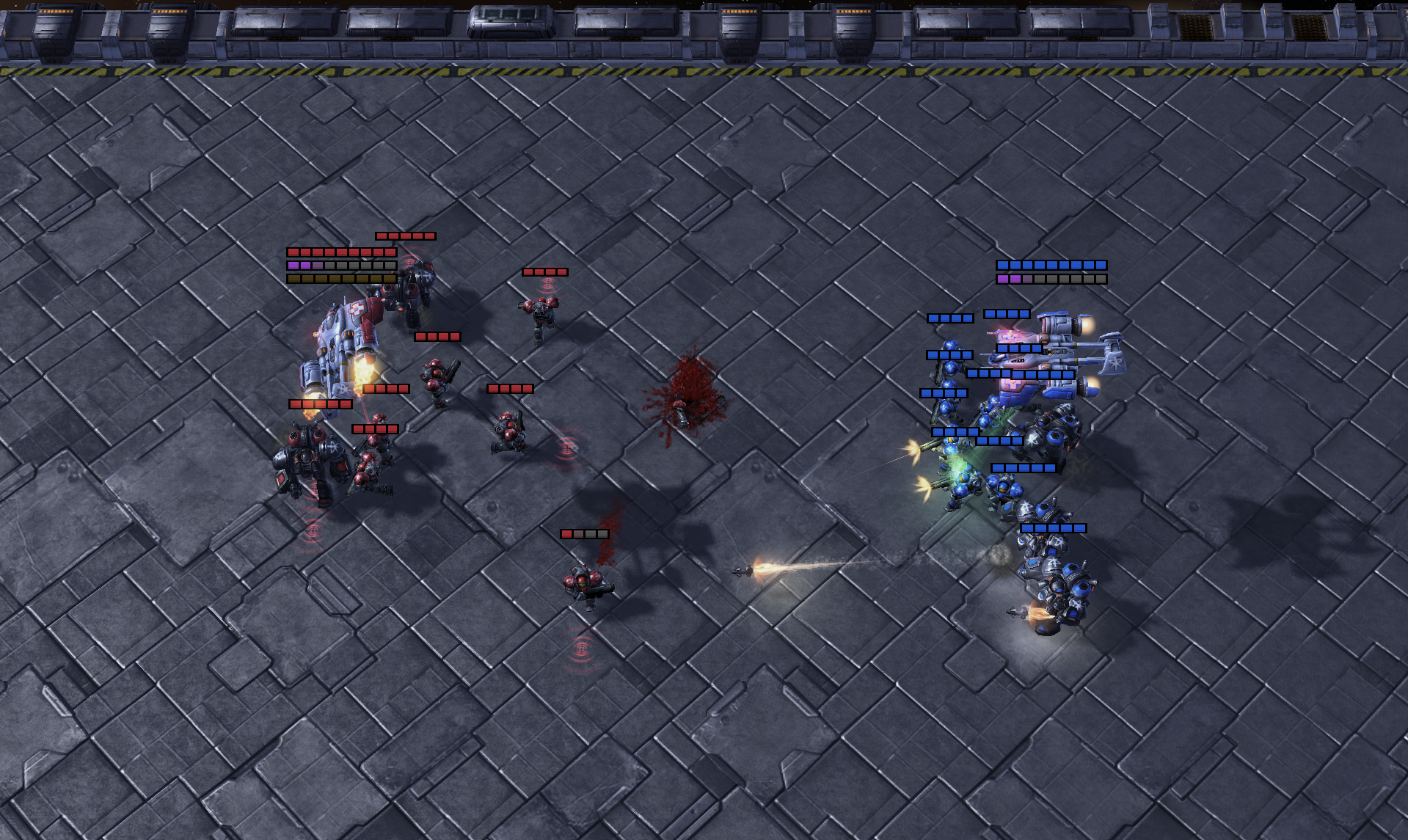}
    \caption{Visualizations of the SMAC environment. \emph{(left)} The \texttt{corridor} scenario. \emph{(right)} The \texttt{MMM2} scenario.}
    \label{fig:smac_screenshots}
\end{figure*}
\subsubsection{Overview}
In this section, we go over the SMAC environment to the extent that is relevant to our project. Our project only replicates SMAC in its default configuration, omitting any optional parameters, so we omit those here as well.

In SMAC, each agent controls one unit (we refer to units controlled by agents as \emph{allied units} throughout this paper) and is tasked with defeating a group of enemy units controlled by Starcraft II's built-in AI opponent. SMAC defines several combat scenarios, differing in army compositions and terrain layout, and as result, in difficulty. We include visualizations of two SMAC scenarios in \cref{fig:smac_screenshots}.

Units are divided into several types with different attributes (health, attack, etc.) -- but note that the SMAC environment only makes a distinction between unit types in the state/observation vectors if there is more than 1 unit type within a single team in the scenario -- we will refer to this as the scenario "distinguishing unit types".

Some unit types, once hit, do not regenerate health in any way. Other units have innate health regeneration that is always active. Yet another group of unit types has special shields on top of their health points that regenerate after a period of not taking damage, and the shields have to be brought down to 0 before the units' health can be hit. Due to the limitations of Starcraft II, either all units in a team possess shields or none of them.

\subsubsection{Actions}
Each agent has access to several actions, which may or may not be available at any given timestep -- the environment exposes a method to get currently available actions for each agent. If an unavailable action is chosen by the agent, SMAC raises an error and ceases execution. The possible actions are \texttt{no-op} -- which has no effect and is only available to dead units, \texttt{stop} -- orders the unit to stop in its place and do nothing, \texttt{moveN}, \texttt{moveE}, \texttt{moveS}, and \texttt{moveW} -- orders the unit to move in the chosen cardinal direction (north, east, south, or west), \texttt{target1}, \texttt{target2}, \ldots -- orders the unit to target the unit with the specified team-specific ID -- for damage-dealing units, this refers to targeting enemy units to attack, for healing units, to targeting allies to heal. SMAC defines a constant targeting range for agents, and this action is unavailable if the target is outside of this range.

\subsubsection{State}
The true state of the environment is a vector divided into three sections. The first section contains each ally unit in order of their IDs: its current health, its current cooldown, its X and Y coordinates (relative to the centre of the map), its current shields (only if allies have shields), and a one-hot vector representing its unit type (only if scenario distinguishes unit types). The second section contains each enemy unit in order of their IDs: its current health, its X and Y coordinates (relative to the centre of the map), its current shields (only if enemies have shields), and a one-hot vector representing its unit type (only if the scenario distinguishes unit types). The final section contains each agent in order of their IDs: a one-hot vector representing the action taken by them in the previous timestep.
Note that all features within the state vector are normalized to be between zero and one -- for example, the health value is divided by the maximum health value of the given unit.

\subsubsection{Observations}
Each agent, in each time-step, receives an observation representing what is visible to the agent within the environment. SMAC defines a constant sight range for agents -- if a unit is dead, or is further from the agents' own unit than this sight range, any information in the observation vector about this unit is completely zeroed out. The observation vector is divided into four sections. The first section includes, for each cardinal direction, whether the movement is possible in that direction. The second section includes, for each enemy unit, if it is alive: whether the agent's unit can attack it, its distance to the agent's unit, its X and Y coordinates relative to the agent's unit, its health, its shields (only if enemies have shields), and a one-hot vector representing its unit type (only if the scenario distinguishes unit types). The third section includes, for each ally unit: a literal 1 (to distinguish from units that are dead or too far), its distance to the agent's unit, its X and Y coordinates relative to the agent's unit, its health, its shields (only if allies have shields), and a one-hot vector representing its unit type (only if the scenario distinguishes unit types). The final section includes, for the agent's own unit: its health, its shields (only if allies have shields), and a one-hot vector representing its unit type (only if the scenario distinguishes unit types). Note that, similarly to the state features, all observation features are normalized to be between zero and one.

\subsubsection{Rewards}
After each time-step all agents receive a shared reward equal to the sum of health points and shield points removed from enemies in that timestep. A small bonus of 10 is added for each eliminated enemy unit and a bigger bonus of 200 is added for winning the scenario. The reward is normalized by dividing it by the sum of all health and shield points of enemy units and any possible bonuses, and multiplying by 20 -- thus, the possible returns should be between zero and twenty. Do note, however, that the actual cumulative reward received by the agents by the end of an episode might exceed 20 due to health and shield regeneration.

\subsection{Optimal Reciprocal Collision Avoidance} \label{sec:orca}
In Starcraft II, units move around the battlefield populated by other units and avoid collision by stepping to the side if they would get in the way of another unit. This makes the battlefield feel more realistic and physical and allows for some advanced strategies like body-blocking or surrounding (in essence, limiting other units' movement by positioning oneself strategically). We felt it was crucial to reproduce this behaviour in our environment. However, because Starcraft II is a proprietary, closed-source piece of software, we cannot use the exact algorithms used in SMAC. To fill this gap we chose the Optimal Reciprocal Collision Avoidance (ORCA) algorithm by \citet{berg_reciprocal_2009}.

The ORCA algorithm fills several criteria desirable for our environment. Much like in SC2, each unit is assumed to be a circle with a specific radius. Units can avoid other units, and they can also avoid static polygonal obstacles. On top of that, the units can not only avoid collisions but also move towards their own goal location at the same time.

The algorithm is parametrized by a time horizon $\tau$. In each run of this algorithm, each unit $A$ computes a set of half-planes (which we call \emph{ORCA half-planes}) in 2D velocity space, each half-plane being the set of velocities safe to choose to avoid collision with some other unit or obstacle $B$ for at least $\tau$ time.

Each unit $A$ considers all units and obstacles in its immediate neighbourhood (i.e. within some radius $r$ around it). For each neighbour $B$, the unit calculates the \textbf{velocity obstacle} induced by the neighbour on it -- that is, the set of positions in velocity space that would make the unit collide (for units: get within the distance of $r_A + r_B$, where $r_A$ and $r_B$ are the units' own radii) with that neighbour within $\tau$ time.

Let $\mathbf{u}$ be the shortest vector from the relative velocity $\mathbf{v}_A - \mathbf{v}_B$ to the velocity obstacle's boundary -- in other words, the smallest amount of change to the relative velocity required to prevent a collision within $\tau$ time. Then, the slope of the line (called the \emph{ORCA line}) defining the ORCA half-plane for that neighbour is given by the slope of the outward normal of the velocity obstacle boundary at point $\mathbf{v}_A - \mathbf{v}_B + \mathbf{u}$. The line is anchored in a point given by $\mathbf{v}_A + \frac{1}{2} \mathbf{u}$. This gives the unit a half-plane of possible velocities that will avoid collision with the neighbour -- note that the adjustment by $\frac{1}{2} \mathbf{u}$ is because the unit assumes the neighbour is following the same algorithm and will adjust by $-\frac{1}{2} \mathbf{u}$ (since from the neighbour's perspective, everything is mirrored -- hence the negation) -- the adjustment is not halved for obstacle neighbours, only for unit neighbours.

Then, given all half-planes induced by neighbours, the algorithm solves a linear programming problem for each unit, to find a velocity that avoids all neighbours and is the closest to the unit's desired velocity. If avoiding collisions completely is not possible, the algorithm solves a different linear programming problem that minimizes the distance the unit crosses behind the ORCA lines.

We go into more detail about how we use this algorithm in SMAClite in \cref{chap:methodology}. Note that this algorithm is not equivalent to a pathfinding algorithm -- when faced with a wall, the units will often stop in front of it, and they will not look very far for a way to go around it.


\section{SMAClite} \label{chap:methodology}
\subsection{Environment implementation}
\begin{figure*}[t]
    \centering
    \includegraphics[trim={7cm 11cm 1.5cm 6cm},clip,height=0.25\textwidth]{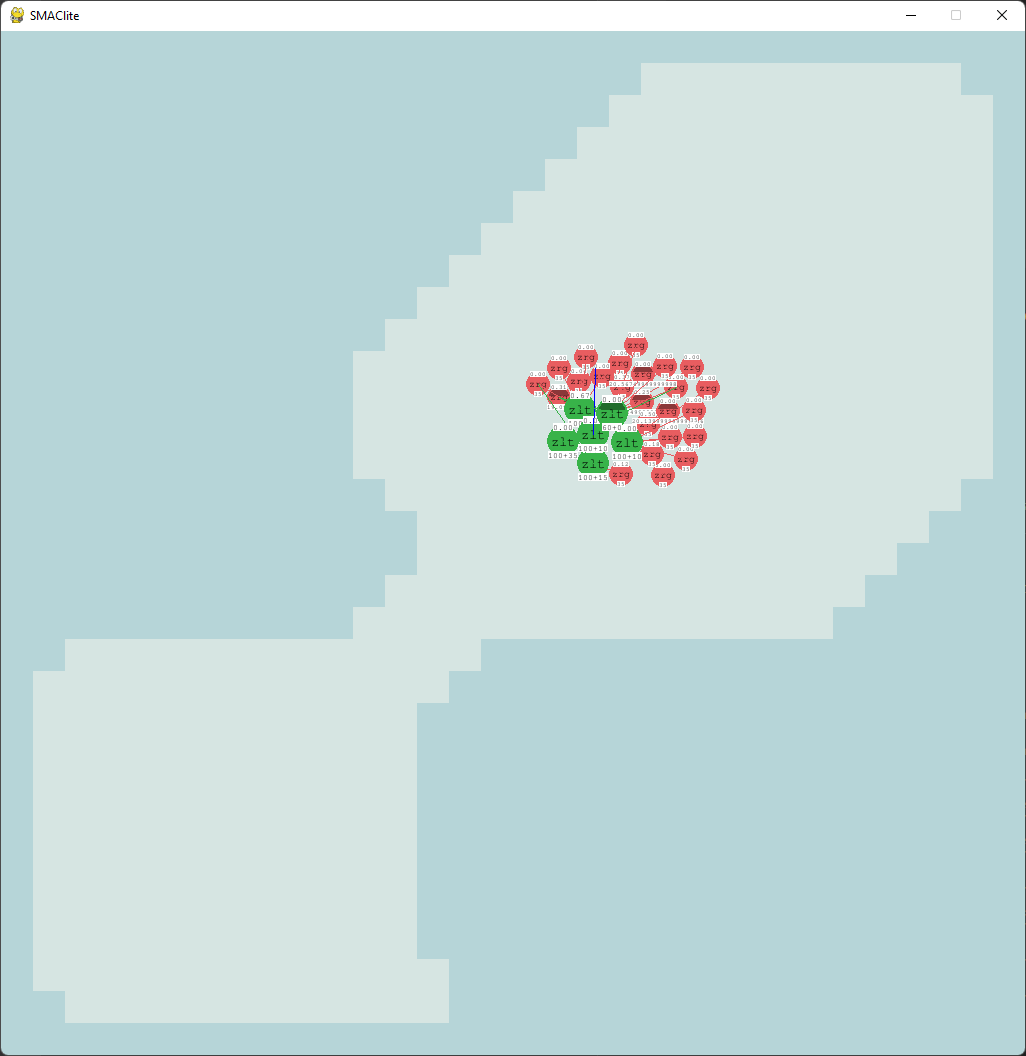}
    \hspace{1cm}
    \includegraphics[trim={1cm 6cm 3cm 7cm},clip,height=0.25\textwidth]{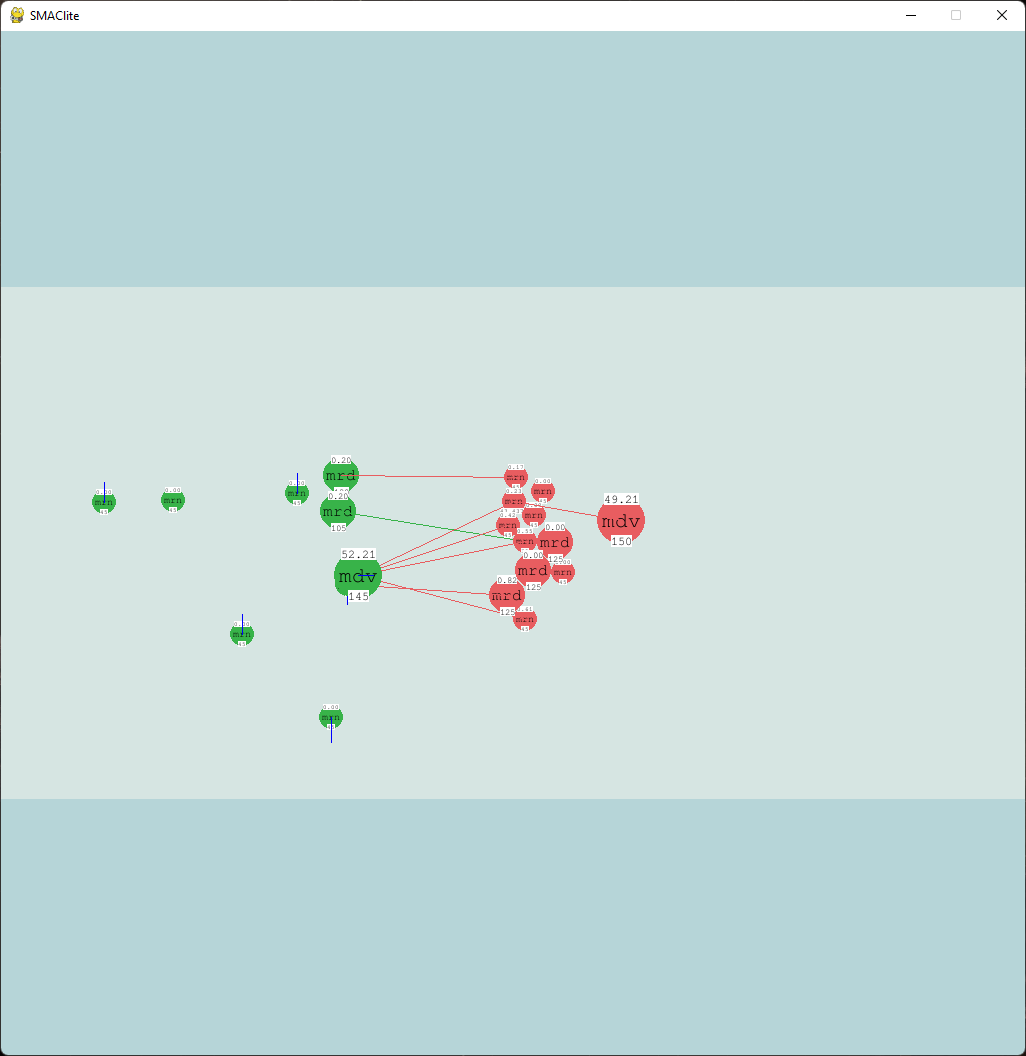}
    \caption{Visualizations of the SMAClite environment. \emph{(left)} The \texttt{corridor} scenario. \emph{(right)} The \texttt{MMM2} scenario.}
    \label{fig:smaclite_screenshots}
\end{figure*}
In this section, we focus on our contribution and how we approached implementing SMAClite. Note that because of the closed-source nature of Starcraft II, we did not have access to any implementation details of it,  algorithms contained in this section were designed with our knowledge of the game and with trial and error experiments, while also using some general information available on the Starcraft II \citet{liquipedia_starcraft_nodate}.

We decided to implement the environment in the Python programming language \cite{van_rossum_python_1995} due to it being widely known in the machine learning community, where it is by far the most popular one. Most of the computations within the environment are performed using the Numpy library \cite{harris_array_2020}, and the rendering of the environment is handled by a script written by us using the Pygame \cite{noauthor_pygame_nodate} library. Wherever applicable, we used the default arguments of the SMAC environment and omitted any optional ones. The SMAClite environment uses the well-established OpenAI Gym framework \cite{brockman_openai_2016} for creating reinforcement learning environments. We include visualizations of two scenarios within the environment in \cref{fig:smaclite_screenshots}\footnote{We also make available videos showing SMAClite in action: \href{https://drive.google.com/drive/folders/1-2YJicUqRzovTa7lRgJilwtecxCPrtVb}{https://drive.google.com/drive/folders/1-2YJicUqRzovTa7lRgJilwtecxCPrtVb}}.

At all steps of the implementation, we made sure the action, state, observation, and reward APIs are exactly aligned with SMAC. This also applies to individual unit attributes, for which we consulted \citet{liquipedia_starcraft_nodate}. This allowed us to conduct transfer learning experiments, such as the one we describe in \cref{sec:zeroshot}.

Even though the map grid in Starcraft II allows triangles within the unitary squares, we chose for simplicity to only allow a grid of squares as the terrain for SMAClite. We found that this simplifying assumption does not detract from the environment's difficulty. Internally, we collapse adjacent squares containing obstacles into rectangles to lower the total number of obstacles for performance's sake. When defining the units' velocities or sizes, the base distance measurement unit is the side length of a single square in the terrain grid. This is consistent with SC2, and all scenarios available in both SMAC and SMAClite use a map size of 32 by 32 units.

To further improve environment performance, the units use K-D trees as available in the Scikit-learn library \cite{pedregosa_scikit-learn_2011} to find their neighbours (e.g. finding units within sight range when generating observation vectors), as opposed to iterating over the entire unit list. Since K-D trees only support querying in a circular area and obstacles are always rectangles, when looking for obstacle neighbours, the units query an R-tree from the Python package \texttt{rtree} \cite{gillies_sean_rtree_nodate}.

\subsection{Base framework} \label{sec:framework}
SMAClite, much like SMAC, is defined mostly by the various combat scenarios available, as well as the units participating in those scenarios. As part of the SMAClite environment, we contribute a framework capable of reading both custom scenarios and units from JSON files -- this means expertise in the Starcraft II map editor is no longer required to create new or modified challenges using the environment. This also means that with SMAClite it is easy to tell the difference between two different unit types -- to compare "zergling" and "marine", all one needs to do is look at their respective JSON definitions, and see what the differences are.

All of the standard scenarios and units shipped with the environment are written using this framework. We give detailed specification of both the scenario and unit definition formats, as well as full examples of JSON files compatible with the framework, in the appendix to this paper.

\subsection{Unit command types} \label{sec:commandtypes}
At any given point in time, each unit in the environment is executing one of several types of commands. SMAClite supports five different command types, with the first four being exactly equivalent to the four action types available to the agents, as described in \cref{sec:smac}.

We introduce one more command type that is unavailable to agents but is key to the AI opponent's behaviour, called \texttt{attack\_move} -- this command orders the units to march toward a specified location, attacking any units encountered along the way, and then guard the location once it is reached. We based our implementation of attack-moving on the "Automatic targeting" article on the Starcraft II \citet{liquipedia_starcraft_nodate}, but made a few judgement calls based on what yielded the most desirable behaviour, wherever their information was unclear or unavailable.

\subsection{Environment loop}
Each environment step starts with the agents' units being assigned commands corresponding to the actions chosen by the agents. Once that happens, we simulate eight game steps, and the reward returned from the environment step is the sum of the rewards earned within these game steps. Each game step is considered to last $\frac{1}{16}$th of a second for the purpose of calculating velocity, cooldowns, etc. -- that does not mean this is its actual duration, as in fact, our environment can run much faster than real-time (see \cref{sec:speed} for details).

Briefly, each game step consists of each unit following the logic associated with its command, e.g. moving, attacking, or waiting. We give a detailed description of each game step in the appendix.

\subsection{Custom ORCA implementation} \label{sec:smacliteorca}
To facilitate installation simplicity, we rewrote the RVO2 library -- the official implementation of ORCA in the C++ programming language from \citet{berg_reciprocal_2009} -- into NumPy, and we ship this module together with SMAClite. Because Starcraft II uses compiled (and, we assume, highly optimized) C/C++ code, and Python code can be quite slow compared to it, we ran into performance issues early in the development of SMAClite. To address this issue, we also make available an addon for SMAClite\footnote{\href{https://github.com/uoe-agents/SMAClite-Python-RVO2}{https://github.com/uoe-agents/SMAClite-Python-RVO2}}, which uses the original C++ RVO2 library verbatim, using Python bindings written in the Cython \cite{behnel_cython_2011} extension. The Python bindings were originally made available by \citet{stuvel_python_nodate}, though we implemented several modifications to suit our use-case -- these modifications are also present in the Numpy version of RVO2. 

The original RVO2 library has no way to remove units one by one or remove all units at once. These are key features for SMAClite, as we need to adjust the collision avoidance unit list whenever a unit dies or whenever we restart the environment. We added both of these features into RVO2 -- we assume dead units disappear from the battlefield as soon as they are eliminated, so we do not want other units avoiding collisions with them.

The second set of adjustments considers static units -- the original ORCA algorithm assumes fully cooperative units that will go out of their way to make the passage easier for other units. Our version, on the other hand, assumes that if a unit $A$ is static (i.e. $||\textbf{v}_A|| = 0$), it will \emph{never} adjust its velocity. This makes it possible to surround other units and/or block their path, which is a valid strategy in Starcraft II. In the original implementation of ORCA, the blocking units would simply be "pushed" away. Note that to make up for this, any moving units will adjust their velocity by $\mathbf{u}$ instead of $\frac{1}{2}\mathbf{u}$ (see \cref{sec:orca} for the definition of $\mathbf{u}$), when avoiding static units -- this ensures moving units will not walk into static units.  

Because the addon requires building and installing C++ files via \texttt{CMake}, which could potentially be problematic on some systems, we chose to make this RVO2 fork available as an optional dependency, for users who are willing to go through a more difficult installation process in order to boost environment performance. We call this version \texttt{SMAClite\_plus}. Note that, because of differences in finding neighbours, unit behaviours will not be exactly the same between the two versions, but should remain functionally indistinguishable. If this addon is in use, custom K-D trees implemented in C++ for RVO2 are used for finding unit and obstacle neighbours for collision avoidance purposes, instead of Scikit-learn K-D trees or \texttt{rtree} R-trees. After it is installed, the addon can be enabled by setting the parameter \texttt{use\_cpp\_rvo2} to \texttt{True} when initialising the environment. Any attempt to set this argument to \texttt{True} without the addon installed will result in an error.

\subsection{Opponent AI behaviour} \label{sec:aiopp}
The authors of SMAC \cite{samvelyan_starcraft_2019} claim the opponent team is controlled by Starcraft II's built-in AI on the \emph{very difficult} level. Because the enemy units' behaviours seemed very simple, we had our doubts about their strategic ability when watching combat inside the SMAC environment. For this reason, we performed the following test on the \texttt{MMM2}, \texttt{2c\_vs\_64zg}, and \texttt{corridor} SMAC scenarios.

First, we toggled the AI level inside SMAC across the various difficulty levels available, while keeping the random number generator seed constant. We then put the opponent AI against agents who pick randomly from the available actions. The opponent units' behaviour was always exactly the same, and we saw no difference at all in the rewards obtained by the random agents between difficulty levels -- they were exactly equal to at least the tenth decimal place. Then, going one step further, we \textbf{removed} the AI opponent from the game, making the enemy units not controlled by any player. This \textbf{did not change the units' behaviour or the resulting rewards either}.

Based on these results, we are reasonably certain Starcraft's built-in AI never issues any orders to the enemy units in SMAC. The only order given to the units is hand-placed inside a script in each SMAC map file -- it tells them to attack-move towards a specific point, usually where allied units initially appear. Following these results, we did not implement any custom opponent AI. All we do is, upon map initialization, set the enemy units' command to \texttt{attack\_move} towards the \texttt{attack\_point} specified in the map scenario file. The enemy units' command never changes throughout the encounter.


\section{Experiments} \label{chap:experiments}
In this section, we describe the results of various experiments we performed in our environment. In all of the experiments, we use all of the scenarios used by \citet{papoudakis_benchmarking_2021}, with the addition of \texttt{bane\_vs\_bane}, which we included to feature a wider selection of unit types. Specifically, we used the \texttt{2sc\_vs\_1sc} (2 stalkers vs 1 spine crawler), \texttt{3s5z} (symmetric map with 3 stalkers and 5 zealots on each side), \texttt{MMM2} (a map with a medivac and some marines and marauders on both sides), \texttt{corridor} (6 zealots vs 24 zerglings in a narrow passageway), \texttt{3s\_vs\_5z} (3 stalkers vs 5 zealots), and \texttt{bane\_vs\_bane} (a map with zeglings and banelings on both sides) scenarios. As optional material helpful to understand each scenario, we attach \cref{tab:scenarios} in the appendix with natural-language descriptions of each scenario.

Through these experiments, we wish to show that our environment does indeed accomplish the goals we set for ourselves. First and foremost, we want to show that SMAClite poses a challenge equivalent to SMAC. We want to show that various MARL algorithms perform in it similarly well as in SMAC, and explain any discrepancies. We also want to prove that SMAClite is strictly cheaper to run than SMAC -- the main metrics we are interested in are the time required to run it, and the RAM it takes up on the machine.
\subsection{Agent learning curves}

\begin{figure*}[t]
    \centering
    \includegraphics[width=\textwidth]{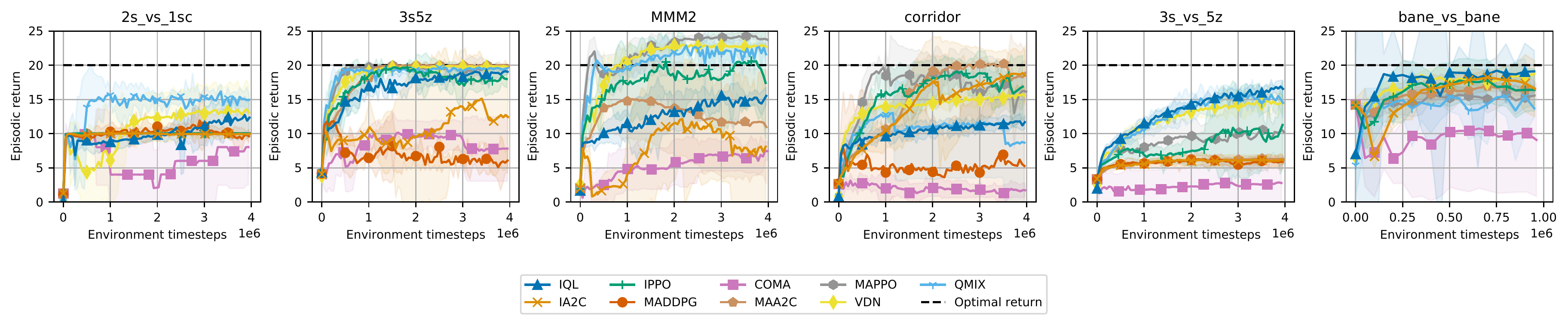}
    \caption{Test-time returns achieved by agents trained using the different algorithms over time during training. The graphs show the mean value, as well as the 95\% confidence interval, from 5 trainings differing by random number generator seed.}
    \label{fig:learning_curves}
\end{figure*}
We trained several MARL algorithms on the selected scenarios. Each training was run for 4 million timesteps, and each training was repeated using 5 different random number generator seeds. All of the training were performed on the \texttt{SMAClite\_plus} version of the environment and were performed solely using CPUs, each training using a single CPU core. The training were run on various nodes on a cloud cluster, most of them using Intel(R) Xeon(R) Gold 6138 CPUs @ 2.00GHz, and all others used CPUs of comparable computational capacity. Due to technical constraints, all training was under a strict 48-hour time limit. For all of the algorithms, we used hyperparameters listed as the best in SMAC as listed in the benchmark paper \cite{papoudakis_benchmarking_2021}. For the episodic algorithms, we used a buffer size of 5000, and for the parallel algorithms, we used a buffer size of 10. We used the EPyMARL framework \cite{papoudakis_benchmarking_2021} to run all of the training.

Some  caveats need to be mentioned with regard to the training procedure. Firstly, the \texttt{bane\_vs\_bane} scenario is by far the slowest (as evidenced by \cref{sec:speed}), so we reduced the number of timesteps to 1 million for this scenario only. In addition, we found the MADDPG algorithm extremely slow during training, and so it never reached the required timesteps in two scenarios (\texttt{MMM2} and \texttt{bane\_vs\_bane}) under the time limit, so we omit those two curves in our figure.

The resulting graphs of episodic return over training time can be found in \cref{fig:learning_curves}. In the remainder of this subsection, we compare the learning curves to those reported for SMAC in the benchmark paper.

First of all, many algorithms easily solve the \texttt{2s\_vs\_1sc} scenario in SMAC, quickly reaching the maximum reward of 20, while in SMAClite the best performer (QMIX) only reaches an average return of about 15. We theorize that this is due to the fact that SMAClite does not simulate attack animations, and originally in Starcraft II the spine crawler has a very long attack animation (0.238 seconds versus 0.1193 seconds for the stalkers), so in SC2 it is much easier to dodge the spine crawler's attack in the last possible moment. Though the difference is less noticeable there, a similar situation occurs in \texttt{3s\_vs\_5z}, and we believe it is for the same reason. Put simply, we believe \emph{kiting} -- alternating between running away and attacking without getting hit -- is much more difficult in SMAClite than in SC2 because of the instant attacks.

Another major difference between the SMAC and SMAClite learning curves is in the \texttt{MMM2} environment -- in the case of SMAClite, three algorithms (MAPPO, VDN and QMIX) seem to have mastered the environment, with maximal rewards almost reaching 25 (for details about how the algorithms achieve this, see the next subsection), while in SMAC the best-performing algorithms barely reach a reward of 17.5 (but note that they are the same three algorithms, which shows the scenarios are still somewhat equivalent). Our hypothesis is that this is due to potentially unintended behaviour in the SMAC environment code. When calculating rewards, SMAC simply subtracts enemy units' new health values from their old health values, so any healing done by the enemies results in a negative reward being incurred by the agents, even though they never did anything wrong. We believe that this, combined with the fact that rewards should be gained for dealing damage, could confuse the agents during training. Judging by the wording in section 4 of the SMAC paper \cite{samvelyan_starcraft_2019} we do not believe this is intentional, so the same behaviour is not present in SMAClite, where we explicitly define the reward as the sum of health points lost by the enemies due to attacks, and we do not penalize the agents for enemies healing.

Despite these few issues, we do observe that the overall shape of the training curves is comparable to those from SMAC in the benchmark paper \cite{papoudakis_benchmarking_2021} in all scenarios -- some of them differ only in maximum return reached. The ranking of algorithms at the end of training time is also largely the same between SMAC and SMAClite. We believe that this is promising evidence pointing towards the environments being equivalent as far as learning is considered.

\subsection{Evaluation of learned behaviours}
In this section, we select one model for each scenario, and we describe the strategies employed by the agents to achieve high reward values, in order to demonstrate the complexity of strategies required in SMAClite. Note that the below descriptions are simply our interpretations of the behaviours demonstrated by the agents in the environment, informed by our knowledge of the game gained both by playing Starcraft II and working on SMAClite.

When making our selection, we wanted to showcase each of the algorithms, and each of them in a scenario where it did well -- note that this means we might not use the best performer in all of the scenarios, but all of them are at least above average. However, because there are 9 algorithms and only 6 scenarios and because not all algorithms performed well, we decided to omit COMA, MADDPG, and IPPO, which were the overall weakest performers. In each case, we used the latest available (i.e. highest amount of training timesteps) checkpoint of agent parameters. \newline
    \texttt{2s\_vs\_1sc} -- QMIX -- \textbf{Mean test return: 16.22} -- one of the stalkers is the "baiter" that gets the spine crawler's attention and they both attack until the baiter's health drops down to a very low value, at which point it backs out and lets the spine crawler target the other stalker. They then both hit the spine crawler until they die, which sometimes results in a victory and sometimes in defeat depending on the random attack order. \newline
    \texttt{3s5z} -- VDN -- \textbf{Mean test return: 20} -- the stalkers and the zealots both focus on the enemy stalkers first, which deal more damage per second than the enemy zealots. The allied zealots move slightly north and the allied stalkers move slightly south, both to direct the attention of enemy stalkers (whose initial position is on the north side of the enemy army) to the allied zealots, who can take more hits, and also to allow allied stalkers plenty of room to maneuver while attacking. Also worth noting is the fact that the allied units are quite good at moving away right as their health drops to a dangerously low level, letting other nearby allies take the aggression. \newline
    \texttt{MMM2} -- MAPPO -- \textbf{Mean test return: 24.63} -- all allied damage-dealers focus on the enemy marauders first, because they are the heavier hitters, while the ally medivac hides behind others as soon as it gets low since it is a priority target and would quickly be gunned down. What is interesting here is that the allied units leave the enemy medivac alive and kill it \emph{last}, abusing the fact that enemy units getting healed results in a higher return due to the total damage dealt is higher -- this is how the agents surpass the optimal return of 20 and reach values close to 25. They even stop hitting enemy damage-dealers and let them get healed, since the medivac cannot heal itself. \newline
    \texttt{corridor} -- MAA2C -- \textbf{Mean test return: 20.25} -- the zealots fan out and form a horizontal line against the zerglings, blocking them making it harder for the zerglings to surround them. This causes the zerglings to crowd around the front of the zealot line, reducing the total amount of damage the zealots take over time. \newline
    \texttt{3s\_vs\_5z} -- IQL -- \textbf{Mean test return: 18.1} -- the stalkers use the intended optimal strategy of kiting the zealots around the map. The lowest-health stalker always makes sure to stand behind the other two when attacking, in order to avoid dying. The stalkers do seem to get "lazy" with their kiting when the number of remaining zealots becomes low, probably because there is no penalty for dying, and it is easier to just stand and attack when the risk of death is low. \newline
    \texttt{bane\_vs\_bane} -- IA2C -- \textbf{Mean test return: 19.14} -- the zerglings run away to the west to avoid the enemy banelings' explosions, while the ally banelings charge forward and explode when they become surrounded by enemy units. The enemy zerglings and banelings quickly die to multiple explosions. If any enemies remain after the allied banelings' explosions, the allied zerglings come out of hiding and attack.

\subsection{Cross-environment zero-shot performance} \label{sec:zeroshot}
\begin{table*}[t]
    \centering
    \caption{Mean test returns achieved by agents trained on SMAClite when put inside the original SMAC environment, achieved using the parameters from the first time they were saved during training, and from the last time they were saved during training. Late return on SMAClite is also included for reference.}
    \begin{tabular}{|c|c|c|c|c|}
    \hline
         Scenario & algorithm & early return & late return & late return on SMAClite \\\hline\hline
         \texttt{2s\_vs\_1sc} & QMIX & 0 & 11.4 & 16.22 \\\hline
         \texttt{3s5z} & VDN & 3.09 & 8.96 & 20 \\\hline
         \texttt{MMM2} & MAPPO & 1.87 & 6.70 & 24.63 \\\hline
         \texttt{corridor} & MAA2C & 3.35 & 4.48 & 20.25 \\\hline
         \texttt{3s\_vs\_5z} & IQL & 3.13 & 9.33 & 18.1 \\\hline
         \texttt{bane\_vs\_bane} & IA2C & 18.97 & 19.84 & 19.14 \\\hline
    \end{tabular}
    \label{tab:zeroshot}
\end{table*}
In order to verify whether our environment does in fact require the same strategic reasoning as SMAC, we performed zero-shot transfer learning experiments on each scenario What we mean by this is, we trained the agents on SMAClite, and then without any retraining put them inside SMAC in the same scenario. We used the same models as in the previous experiment -- to be specific, we first used their parameters from the first time the model was saved (at the very beginning of the training process), and then their parameters from the last time the model was saved (at the very end of the training process), and then compared the mean test-time returns. We present the results in \cref{tab:zeroshot}.

We note that the mean return obtained has increased in \emph{all} of the scenarios, with some exhibiting significant improvements such as doubling or tripling of the mean return. A noteworthy example is the \texttt{bane\_vs\_bane} scenario, where even the early version of the agents got close to the optimal return of 20, but was still improved upon by the later version. Because all of the scenarios exhibited improved returns upon training in SMAClite, we believe there is evidence to support the two environments requiring similar sets of skills, and that transfer learning from one to the other is a viable training strategy.

\subsection{Environment Performance Benchmark}  \label{sec:speed}
In this section, we consider the performance of the environments themselves, to confirm that SMAClite is indeed cheaper to run than SMAC. To obtain the data in this benchmark, we ran each scenario 20 times with agents picking randomly among the available actions. All of the below experiments were run on a computer with an AMD Ryzen 3700X CPU.

\begin{table*}[t]
    \centering
    \caption{Average seconds per timestep on the SMAC scenarios we used for training. Data was obtained by running the scenario 20 times against random agents.}
    \begin{tabular}{|c|c|c|c|c|c|}
        \hline
         Scenario & SMAC & SMAClite & Change & SMAClite\_plus & Change \\\hline\hline
         \texttt{2s\_vs\_1sc} & 0.004 & 0.007 & +75\% & 0.003 & -25\% \\\hline
         \texttt{3s5z} & 0.013 & 0.018 & +38\% & 0.007 & -46\% \\\hline
         \texttt{MMM2} & 0.017 & 0.028 & +64\% &  0.010 & -41\% \\\hline
         \texttt{corridor} & 0.014 & 0.092 & +557\% & 0.010 & -28\% \\\hline
         \texttt{3s\_vs\_5z} & 0.006 & 0.013 & +116\% & 0.005 & -17\% \\\hline
         \texttt{bane\_vs\_bane} & 0.049 & 0.086 & +76\% & 0.024 & -51\% \\\hline
    \end{tabular}
    \label{tab:speed}
\end{table*}

In \cref{tab:speed} one can find the time each environment took per timestep (excluding logic not belonging to the environment, like action selection). We notice that the pure Python code is indeed slower than the original SMAC environment, but the environment becomes much faster when using the C++ RVO2 addon.

Note that these timings correspond to running a single environment step, which consists of 8 game steps in sequence, together with any environment-only logic (e.g. calculating rewards and determining observations). If we ever were to run experiments against human players (assuming SMAClite got some human control extension), we would want the environment to be capable of running in real-time. Both in SC2 and SMAClite, each game step, of which there are 8 in an environment step, is considered to last $\frac{1}{16}$ seconds, and most human players play SC2 at the "faster" in-game speed, which corresponds to a 40\% speed-up \cite{liquipedia_starcraft_nodate}. Therefore, in order for the environments to run in real-time, they can afford to use $\frac{8}{16 * 1.4} \simeq{0.357}$ seconds per environment step. Thus, all versions of the environment are more than capable of running all tested scenarios in real-time -- the advantage of \texttt{SMAClite\_plus} becomes the most obvious when running lengthy training which requires many executions of the environment. 

While running the experiments described above, we also measured the amount of RAM used by each environment. This did not vary a lot by scenario and oscillated around 600 MB for the \texttt{SMAC} environment, and around 100 MB for the \texttt{SMAClite} and \texttt{SMAClite\_plus} environments. Therefore, our lightweight version of the environment requires six times less memory than the original SMAC environment to run.


\section{Future Work And Conclusion}
We presented SMAClite -- a lightweight environment for MARL, consisting of a game engine emulating the Starcraft II minigame of SMAC, as well as a framework for easily creating new scenarios and units for this engine, using a familiar JSON format. We conducted experiments to show that SMAClite presents a challenge equivalent to SMAC, both by comparing learning curves and MARL algorithm rankings, as well as through a zero-shot learning experiment where training on SMAClite improved the agents' SMAC performance. We also showed that this challenge comes at a much-reduced cost, both in terms of required time and memory.

The SMAClite engine and framework are both very much open to extensions. Thanks to the fact that the environment is no longer bound by the Starcraft II dependency, developers could introduce game mechanics unrelated to Starcraft II, and could also go beyond the game's technical limitations. One could implement new unit types unseen in SC2 with new attack types or abilities, or easily create countless intricate puzzles for the agents to solve using the scenario framework.

It is also possible to treat SMAClite strictly as an extension of SMAC and work towards making it as close to SMAC as possible while maintaining the performance improvements it brings. This would likely mean focusing on transfer learning experiments such as the one in \cref{sec:zeroshot}, and eliminating various differences present currently between the environments. One example might be a projectile/animation simulation, which is one of the major differences between the SMAC and SMAClite, and caused discrepancies in \texttt{2s\_vs\_1sc}, as well as in \texttt{3s\_vs\_5z}, during our experiments.

One noteworthy contribution to the SMAC ecosystem is SMAC v2 \cite{ellis_smacv2_2022}, which introduces procedurally generated scenarios that change episode-to-episode, and show that this stochasticity makes for more challenging scenarios and forces the agents' strategy to be more adaptable. This is certainly in interesting direction, and we will look into implementing similar improvements to SMAClite in the nearest future.

One could also work on improving the modified ORCA algorithm's combat capabilities -- as evidenced by the \texttt{corridor} scenario, the SC2 zerglings were much better at surrounding the zealot wall as performed by our MAA2C agents than the SMAClite zerglings. This probably stems from the fact that RVO2 is mostly a general-purpose collision avoidance algorithm, while SC2's algorithm was handwritten for the best combat performance possible. Another feature idea we could use from SC2 is a pathfinding algorithm, since running in a straight line becomes a problem very soon when the terrain becomes any more complicated than the terrain in our experimental scenarios.

\newpage
\bibliographystyle{ACM-Reference-Format} 
\bibliography{references}

\clearpage
\begin{appendices}
\begin{table*}[t]
    \caption{Combat scenarios used in our experiments.}
    \centering
    \begin{tabularx}{\textwidth}{|c|c|c|X|}
    \hline
         Name & Allied Units & Enemy Units & Description  \\
         \hline \hline
         \texttt{2s\_vs\_1sc} & 2 & 1 & Two stalkers -- powerful but fragile ranged units -- face off against one spine crawler -- very strong unit with no movement capabilities. The stalkers need to abuse the spine crawler's immobility to bring its health down. \\\hline
         \texttt{3s5z} & 8 & 8 & A symmetrical map with each team having three stalkers and five zealots -- melee units that deal less damage but can take a lot of hits before dying.\\\hline
         \texttt{MMM2} & 10 & 12 & Each team has some marines -- fragile ranged units that attack quickly, some marauders -- more durable units that have stronger attacks, but can't hit flying units, and one medivac -- a flying healer unit. \\\hline
         \texttt{corridor} & 6 & 24 & The allied team only has 6 zealots to hold off 24 zeglings -- extremely fragile but very quick units with potential to overwhelm unprepared enemies with numbers and surround them.\\\hline
         \texttt{3s\_vs\_5z} & 3 & 5 & Three stalkers need to keep the quick zealots at bay while outnumbered, using their range to their advantage.\\\hline
         \texttt{bane\_vs\_bane} & 24 & 24 & Each team has some zerglings and some explosive kamikaze banelings, each of which can easily take out several zegrlings with one explosion.\\\hline
    \end{tabularx}
    \label{tab:scenarios}
\end{table*}
\section{SMAClite framework details}
\subsection{Scenario definition}
Our framework accepts scenario JSON files containing a single JSON object. Each scenario should have a \texttt{name} to identify it, and should specify the numbers of allied and enemy units with the \texttt{num\_allied\_units} and \texttt{num\_enemy\_units} parameters. The units in the scenario should be listed with the \texttt{groups} parameter, each group being a JSON object with \texttt{x} and \texttt{y} parameters specifying the group's center, a \texttt{faction} parameter (\texttt{ALLY} or \texttt{ENEMY}) to specify which team the units are on, and a \texttt{units} parameter with an object of unit types together with their counts. Each group will initially be laid out in the shape of a square around their specified location. Note that these groups have no impact beyond unit positioning -- once initial unit placement is complete, neither the agents nor the units have any information about what group they came from.

The framework supports two ways of specifying unit types. In order to use a standard unit type, its uppcercase name should be used (e.g. \texttt{ZERGLING}). The other way to specify a unit type is to provide a path to a JSON file with its specification. The framework also supports an optional \texttt{custom\_unit\_path} parameter -- if specified, this path will be prepended to all unit types specified as a path. Note that if the \texttt{.json} extension is missing the framework reattaches it automatically, so if the custom unit path is \texttt{path/to} and the unit type specification is \texttt{type}, the framework will look for the file \texttt{custom/unit/type.json} file. The framework also requires an \texttt{attack\_point}, which is typically near to the initial positions of the allied units -- this point is where the enemy units will be marching towards throughout the scenario (see \cref{sec:aiopp} for details).

The framework also supports two ways of defining terrain -- one way is to pick a standard terrain present by supplying the \texttt{terrain\_preset} parameter with its uppercase name (e.g. \texttt{CORRIDOR}). Terrain can also be provided in the scenario definition file itself using the \texttt{terrain} argument, which should consist of a list of strings forming a rectangular 2D array. The framework supports two types of terrain: \texttt{\_} for walkable, and \texttt{X} for non-walkable. The framework also always requires a \texttt{width} and \texttt{height} to be specified for the scenario, which should match the terrain dimensions -- note that all scenarios adapted from SMAC have the dimensions of 32 by 32.

Finally, the framework requires some general details about units participating in the scenario. The parameters \texttt{ally\_has\_shields} and \texttt{enemy\_has\_shields} specify which teams, if any, should have shields active on them, and the parameters \texttt{num\_unit\_types} and \texttt{unit\_type\_ids} specify what IDs the agents will receive in observations for the various unit types. The former should be a number (note that scenarios adapted from SMAC use 0 for all scenarios where both teams are homogenous, i.e. only have one unit type each), and the latter should be a map from unit type specifications (same as above) to numbers from 0 to \texttt{num\_unit\_types} minus one. The length of the map must be equal to \texttt{num\_unit\_types}.

We ship several standard scenarios with this environment as separate OpenAI Gym environments (e.g. \texttt{smaclite/2s3z-v0}). To use a custom scenario file, one should use the \texttt{smaclite/custom-v0} Gym environment, and provide a path to the scenario file via the \texttt{map\_file} parameter.

\subsection{Unit definition}
Each unit is assigned a specific \textbf{type}, with a set of different attributes impacting the environment mechanics. Our framework supports several attributes in the JSON files defining the various unit types, in order to allow for their easy customization.

Firstly, the framework supports several attributes defining the units' resources, including their maximum health and health regeneration via the \texttt{hp} and \texttt{hp\_regen} attributes, respectively. Their shields, if any, via the \texttt{shield} attribute, and their maximum and initial energy via the \texttt{energy} and \texttt{initial\_energy} attributes. The units' size can be specified by providing their diameter in the \texttt{size} attribute, and their speed (in distance per second) via the \texttt{speed} parameter.

Then, the framework supports several attributes defining the units' combat abilities. Their \texttt{combat\_type} (always one of \texttt{DAMAGE} or \texttt{HEALING}) defines their main role on the battlefield, and their \texttt{damage} defines how strong of a hitter they are, while their \texttt{armor} defines their defensive capabilities. Each unit has an \texttt{attack\_range}, which defines how close (measured boundary to boundary) the unit has to get to a target in order to attack or heal it\footnote{Note that this is different from the agent targeting range mentioned in \cref{sec:smac} -- that range only affects the agents' observations, while this one actually governs when an attack can happen.}. Note that while a numeric value is usually expected for this attribute, the special value of \texttt{MELEE} is also accepted to signify that the unit should have the standard melee range, i.e. only attack from up close. Some units can deal damage multiple times per attack -- this can be achieved using the \texttt{attacks} attribute (e.g. 2 for attacking twice at once), and after attacking, each unit has to wait their specific \texttt{cooldown} before attacking again. Each unit type should also define a \texttt{minimum\_scan\_range}, which will govern how far the units will look when searching for targets.

In SMAClite, each unit resides in a specific plane, with three currently supported: \texttt{GROUND}, \texttt{AIR}, and \texttt{COLOSSUS}. When moving, the units only avoid collisions with units in the same plane as them, and only ground units are affected by static obstacles. All of the units in SMAClite can only target units which reside in the planes listed in their \texttt{valid\_targets} -- for example, if \texttt{AIR} is not in this list for some unit, then it can never attack airborne units. But note that, regardless of their \texttt{valid\_targets}, all units can target units int he \texttt{COLOSSUS} plane.

In order to support making certain unit types stronger against specific other types, the framework also supports a system of unit \texttt{attributes} (e.g. \texttt{BIOLOGICAL}) and attribute-relative \texttt{bonuses}, defined as a map from attribute to bonus value. For example, if a unit type has a bonus of 20 against \texttt{ARMORED} units, it will deal 20 bonus damage with each attack against units with that attribute. 

Finally, the framework supports several different attack types for units, defined by the \texttt{targeter} and \texttt{targeter\_kwargs} attributes. The most common attack type is \texttt{STANDARD}, which simply has the unit attacking one target at a time. The other attack types are only used by one standard unit each, but can easily be reused for custom unit types. The \texttt{KAMIKAZE} attack type has the unit explode when attacking, dealing damage in a circle with a specified \texttt{radius} around itself and dying the process. The \texttt{LASER\_BEAM} type fires a laser in a line perpendicular to the line between the attacker and their target -- more specifically, the laser line is a rectangle with a specified \texttt{width} and \texttt{height}. Lastly, the \texttt{HEAL} attack type is used by healer units.

\section{SMAClite framework examples}

\subsection{Example of a valid scenario file} \label{chap:exampplescenario}
The following is an example custom scenario similar to the built-in scenario \texttt{10m\_vs\_11m}, but using custom units.
\begin{verbatim}

{
    "name": "10m_vs_11m",
    "custom_unit_path": "smaclite/env/units/smaclite_units",
    "num_allied_units": 10,
    "num_enemy_units": 11,
    "groups": [
        {
            "x": 9,
            "y": 16,
            "faction": "ALLY",
            "units": {
                "example_custom_unit": 10
            }
        },
        {
            "x": 23,
            "y": 16,
            "faction": "ENEMY",
            "units": {
                "example_custom_unit": 11
            }
        }
    ],
    "attack_point": [9, 16],
    "terrain_preset": "NARROW",
    "num_unit_types": 0,
    "ally_has_shields": false,
    "enemy_has_shields": false
}
\end{verbatim}
\subsection{Exmaple of a valid unit file} \label{chap:exampplescenariounit}
The following is a custom unit similar to the built-in \texttt{MARINE} unit, but with a much larger (effectively global) scan range
\begin{verbatim}
{
    "hp": 45,
    "armor": 0,
    "damage": 6,
    "cooldown": 3,
    "speed": 3.15,
    "attack_range": 3,
    "size": 3,
    "attributes": ["LIGHT", "BIOLOGICAL"],
    "minimum_scan_range": 100,
    "valid_targets": ["GROUND", "AIR"]
}
\end{verbatim}

\section{Details on the SMAClite Game loop}
In this section we describe in detail the logic of each game step, of which there are several within each environment step. Each game step consists of several phases. These phases were developed by us and we have no information about whether the game steps inside Starcraft II follow any sort of similar order -- this structure was simply what yielded the most reasonable game ruleset that is resembles Starcraft II. Before the next phase can begin, all units must execute the logic for the previous phase. This is necessary to ensure the units' perceptions of other units (note: unrelated to the agents' observations) remain consistent throughout the execution of the game step.

\subsection{Target clean-up}
First of all, each unit might lose the target it was attacking or healing in the previous game-step, according to command-specific logic. It is necessary for all units to execute this logic before proceeding with declaring their preferred velocity for this game step because other units might access the target information when computing their own preferred velocity.

Units with the \texttt{noop}, \texttt{stop}, and \texttt{move} commands immediately lose any target they had, since these are non-combat commands. Units with the \texttt{target} command acquire the target dictated by their command, or retain it they are already targeting it. With the \texttt{attack\_move} command, the units consider several factors. They always lose a dead target and they never lose a target who attacked them in the last game step. Otherwise, they lose their target if it is outside of their attack range.

\subsection{Velocity preparation}
In this phase all units declare their preferred velocity for this game step. It is necessary that they all do this before proceeding with velocity adjustment via the ORCA algorithm, because units can perceive each other's preferred velocity and make collision avoidance decisions based on that information.

In the case of \texttt{noop} and \texttt{stop} commands, the preferred velocity of the units is always $\mathbf{0}$. In the case of the \texttt{move} command, if the unit's maximum velocity is $v_{max}$, its current position is $\mathbf{x}$, and the position where it wants to move is $\mathbf{y}$, where $\mathbf{x} \neq \mathbf{y}$, then the unit always declares that its preferred velocity is $(\mathbf{y} - \mathbf{x}) \frac{v_{max}}{||\mathbf{y} - \mathbf{x}||}$. If $\mathbf{x} = \mathbf{y}$, then the preferred velocity is instead $\mathbf{0}$. Put simply, the unit always wishes to move in a straight line towards its destination -- this is different to Starcraft II's engine which has a built-in pathfinding algorithm based on the A* algorithm \cite{hart_formal_1968} -- but we decided to omit it, since it only matters in one of the many scenarios offered by SMAC (\texttt{2c\_vs\_64zg}).

If the unit's command is \texttt{target}, the unit considers its distance from the target unit. Let the unit's attack range be $d_{max}$, its radius $r_A$, and its target radius $r_B$, and let the distance between it and the target be defined as $d(A, B)$. Then, in the case where $d(A, B) > r_A + d_{max} + r_B$, the unit declares it is moving in a straight line towards its target, and proceeds as with a \texttt{move} command. Otherwise, it declares it is attacking or healing, and its preferred velocity is $\mathbf{0}$.

Finally, if the unit's command is \texttt{attack\_move}, the unit first finds the valid targets within its scan range. It is important to note that if the unit is a damage-dealer, it considers any enemy healers \emph{priority targets}, and all other enemy units as non-priority targets -- if the unit still has a target after the target clean-up phase, the only way it will switch targets at this point is if its current target is not a priority target, and there is a priority target within its scan range. If it does not have a target, or needs to switch, it picks the closest target with the highest available priority among the valid targets. The process is slightly different with healers, who consider valid targets any non-healer units in their team who are below full health, or who are attacking another unit. The healer unit picks as its target the lowest-health unit among its valid targets. After the target selection process finishes, the unit proceeds as with a \texttt{target} command if it has a target, and as with a \texttt{move} command toward its attack-move destination position if it does not. Note that healers will adjust their maximum velocity for any given game step to match the slowest allied unit within their scan range -- we implemented this behaviour in order to stop them getting in front of their army if they are the fastest units.

There are a few intentional changes from SC2 in this phase -- therein, it is not true that healers can target any non-healers, and can never target other healers or themselves. This is only true with the limited set of units present in SMAC, and we felt it makes for a nice simplification of the SC2 ruleset that does not change anything in SMAC. It is also not true that damage-dealers consider any healers as priority targets in SC2 -- they only do in SMAC due to the modifications made to the map files by its authors -- again, we felt this is a nice simplification aligned with SMAC, but not with SC2.

\subsection{Velocity adjustment}
Each unit $A$ uses the ORCA algorithm to determine its actual velocity that avoids collisions, taking into account its own preferred velocity, as well as its neighbour units and obstacles. For the purpose of collision avoidance, the unit considers its neighbours all units within the radius of $(r_A + r_{max})\tau$ of itself, where $r_{max}$ is the maximum radius of any unit in the scenario and $\tau$ is the time horizon; and all obstacles within the radius of $r_A + \tau v_{A,max}$ of itself. In SMAClite, we always use $\tau = 1$, i.e. the units want to guarantee avoiding collisions for 1 second, or 16 game steps. The ORCA algorithm returns for each unit its actual velocity, given its preferred velocity. For details on our implementation of the ORCA algorithm, refer to \cref{sec:smacliteorca}.

\subsection{Game step execution}
This is the final phase of the game step, and because order matters here (a unit might be eliminated before it gets to execute its command), the units execute their respective logic for this phase \emph{in random order}, controlled by the environment randomness seed, with the order being re-randomized in each game step. Note that we do not have any data about whether Starcraft II also randomizes game step execution order, this simply felt like a decent compromise between fairness and simplicity.

Each unit, regardless of its command type, begins this phase by performing several standard updates. It updates its position according to its actual velocity computed in the previous phase (i.e. $\mathbf{x}_{new} = \mathbf{x} + \frac{\mathbf{v}_{actual}}{16}$, since velocity is always defined in distance per second), it reduces its cooldown if applicable (i.e. if it is greater than 0, it gets reduced by $\frac{1}{16}$th of a second), and it regenerates health, energy, and/or shields, if applicable. It then proceeds with command-specific logic.

The only case where any command-specific logic is performed at this point is if the unit's command is \texttt{target}, or \texttt{attack\_move} with a non-empty target, and the unit declared during the velocity preparation phase that it is attacking or healing, and its current cooldown is 0 -- in all other cases the unit's game step logic ends here. In the single actionable case, it attacks or heals its target.

Damage from attacks is dealt first to the enemy shields, then to the enemy health, with any damage dealt to the target's health being reduced by its armor. If the unit's \texttt{attacks} attribute is greater than 1, then it deals damage multiple times in a row during this step. If the unit did attack, its cooldown is then set to its maximum value, defined by the unit's type. Healing units heal their target at a rate of 9 health per second, spending $\frac{1}{3}$ of an energy point per health healed.

Do note that units might attack or heal units that are technically outside of their range at this point, because attack/heal declaration happened during the velocity preparation phase, and the target might have moved away slightly since then. This is intentional and is meant to prevent endless chases when the units' velocities are similar. Again, we do not have data on whether the Starcraft II engine features any similar simplifications, but this rule works well for SMAClite.

In addition, there is an intentional change from SC2 in here as well, because for simplicity we do not simulate attack animations or attack projectiles -- every attack happens instantly in the timestep when the unit declares it is attacking. This causes some small discrepancies from SC2, but we believe the margin of error is acceptable.
\end{appendices}


\end{document}